\documentclass[letterpaper, 10 pt, conference]{ieeeconf}
\IEEEoverridecommandlockouts 
\usepackage{amsmath,amsfonts,amssymb, bm}
\usepackage{cite}
\usepackage{hyperref}
\usepackage[pdftex]{graphicx}
\usepackage{epstopdf}
\usepackage{array}
\usepackage{wasysym}
\usepackage{bm}
\usepackage[per-mode=symbol]{siunitx}
\usepackage{url}
\usepackage{xcolor}
\usepackage{epstopdf}
\usepackage[ruled]{algorithm2e}
\usepackage{multirow}
\usepackage{booktabs}
\usepackage{lineno}
\usepackage{bm}
\usepackage{mathrsfs}
\usepackage{upgreek}
\usepackage{tabularx} 

\usepackage{threeparttable}

\title{\LARGE \bf
Event Camera-based Visual Odometry for Dynamic Motion Tracking of a Legged Robot Using Adaptive Time Surface}

\author{Shifan Zhu$^{1}$, Zhipeng Tang$^{1}$, Michael Yang$^{1}$, Erik Learned-Miller$^{1}$, and Donghyun Kim$^{1}$% <-this % stops a space
\thanks{$^{1}$University of Massachusetts Amherst, 140 Governors Dr, U.S. {\tt\small donghyunkim@cs.umass.edu}}
}

\begin{document}

\maketitle
\thispagestyle{empty}
\pagestyle{empty}

%%%%%%%%%%%%%%%%%%%%%%%%%%%%%%%%%%%%%%%%%%%%%%%%%%%%%%%%%%%%%%%%%%%%%%%%%%%%%%%%
\begin{abstract}
Our paper proposes a direct sparse visual odometry method that combines event and RGB-D data to estimate the pose of agile-legged robots during dynamic locomotion and acrobatic behaviors. Event cameras offer high temporal resolution and dynamic range, which can eliminate the issue of blurred RGB images during fast movements. This unique strength holds a potential for accurate pose estimation of agile-legged robots, which has been a challenging problem to tackle. Our framework leverages the benefits of both RGB-D and event cameras to achieve robust and accurate pose estimation, even during dynamic maneuvers such as jumping and landing a quadruped robot, the Mini-Cheetah. Our major contributions are threefold: Firstly, we introduce an adaptive time surface (ATS) method that addresses the whiteout and blackout issue in conventional time surfaces by formulating pixel-wise decay rates based on scene complexity and motion speed. Secondly, we develop an effective pixel selection method that directly samples from event data and applies sample filtering through ATS, enabling us to pick pixels on distinct features. Lastly, we propose a nonlinear pose optimization formula that simultaneously performs 3D-2D alignment on both RGB-based and event-based maps and images, allowing the algorithm to fully exploit the benefits of both data streams. We extensively evaluate the performance of our framework on both public datasets and our own quadruped robot dataset, demonstrating its effectiveness in accurately estimating the pose of agile robots during dynamic movements.

Supplemental video: \url{https://youtu.be/-5ieQSh0g3M}
\end{abstract}
%%%%%%%%%%%%%%%%%%%%%%%%%%%%%%%%%%%%%%%%%%%%%%%%%%%%%%%%%%%%%%%%%%%%%%%%%%%%%%%%

\section{INTRODUCTION}
Legged robots are developed to tackle a range of demanding tasks, such as disaster response~\cite{yoshiike2017development}, search-and-rescue operations~\cite{lindqvist2022multimodality} \cite{delmerico2019current}, patrolling and exploring challenging environments such as forests, mountains, underwater, and even space~\cite{ha2020quadrupedal,miki2022learning,kim2020vision,zhang2022vision,arm2019spacebok,picardi2020bioinspired}. To navigate through such rough terrain, one essential function is to accurately estimate a robot's position and orientation. However, the accuracy of traditional RGB-based visual odometry (VO) or integration of VO and inertia measurement unit (VIO)~\cite{yousif2015overview, aqel2016review, qin2018vins} can significantly drop in dark or highly dynamic environments, where images can be blurred or under-exposed. To address this issue, researchers began to utilize a new type of camera, called event cameras, which offer microsecond-scale temporal resolution and a high dynamic range of up to $140~\si{\decibel}$, compared to the standard RGB camera's dynamic range of $60~\si{\decibel}$.

Event cameras differ from RGB cameras in that they detect brightness changes in the scene asynchronously and independently for every pixel, which allows for almost continuous sensing without image blur even under dynamic movement or low-light conditions~\cite{survey2020event}. Unlike RGB cameras which generate images with 3-channel values, the event data stream includes the position, timestamp, and polarity of emitted events. A common approach to using event data is constructing an event image by accumulating events in a single frame at a constant rate, then applying techniques developed for RGB images. \cite{kueng2016low,vidal2018ultimate,guan2022pl} proposed feature-based algorithms, which detect and track features (e.g. corners, lines, etc) from event images. Although the prior works showed impressive tracking performance during dynamic movements, the feature detection and tracking are not robust to random movements because the event image depends on not only texture but also the motion of the camera. This motion-dependency problem is particularly significant in legged robots, which involve sudden motion direction changes and impact disturbances from touch-downs and jumps~\cite{dhkim2020vision}. In addition to these feature detection and tracking challenges, the process of feature extraction and matching is time-consuming and may require sacrificing the low-latency features of an event camera.

\begin{figure}
    \centering
 \includegraphics[width=\linewidth]{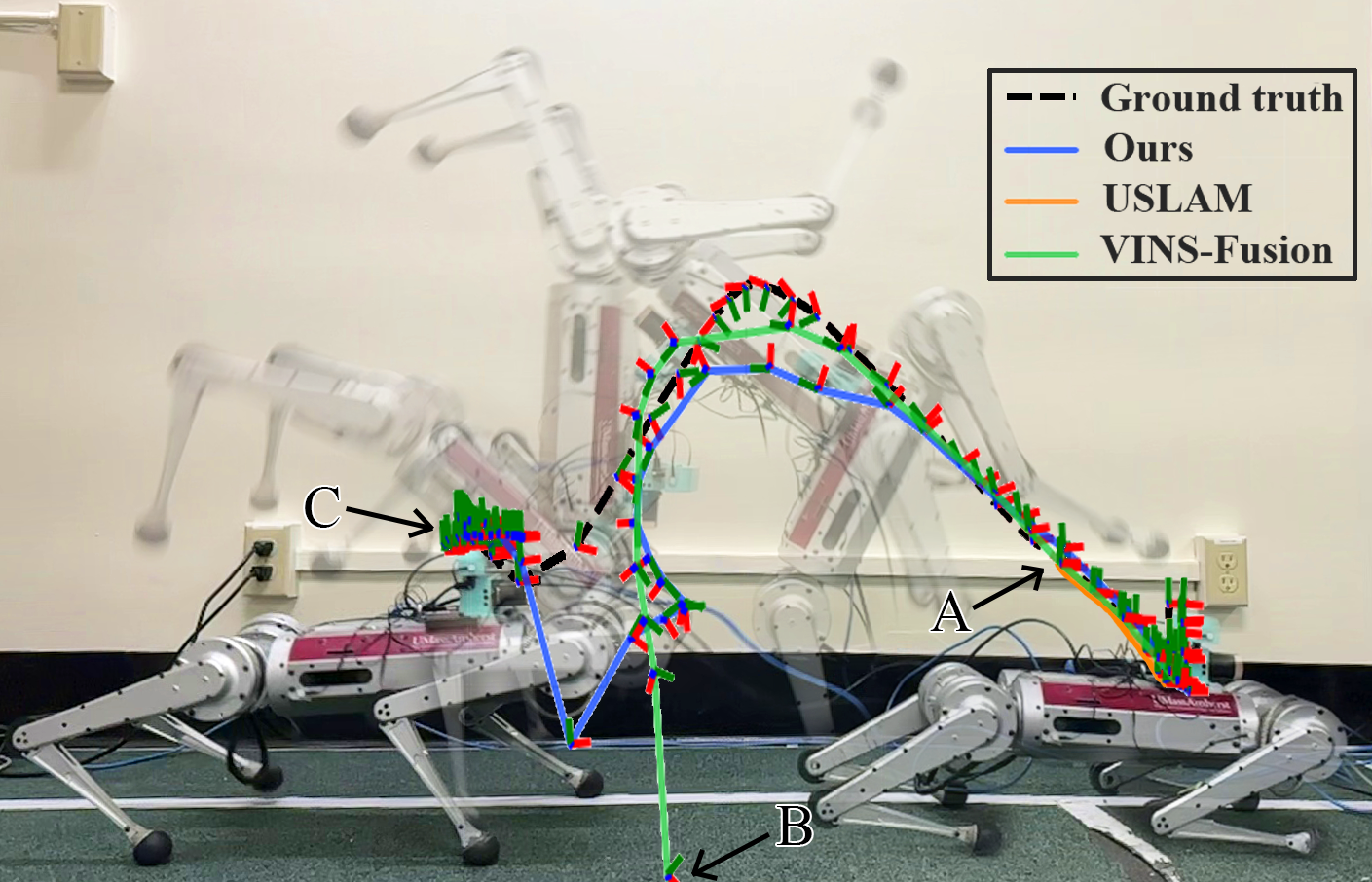}
    \caption{\textbf{Pose estimation during the backflip of Mini-Cheetah.} The trajectory of the robot is shown in transparent figures, while the trajectories from different algorithms are shown in different colors. USLAM~\cite{vidal2018ultimate} terminates at point A. VINS-Fusion~\cite{qin2019a} diverges at point B. Our method successfully converges to the accurate position C.}
    \label{fig:backflip}
\end{figure}

Contrary to feature-based (indirect) methods, direct methods~\cite{engel2017direct} use pixel brightness to estimate the pose, bypassing the need to identify and match features. By using abundant information from images, direct methods are more robust and accurate as long as the environment brightness is consistent and the image gradient transitions gradually. However, event images are constructed by binary event data, resulting in discrete changes in gradients that make the optimization difficult to converge smoothly to a correct pose. A popular approach to remedy the issue is to use a time surface~\cite{hots2017TS}, which is a 2D map constructed by decaying grayscale values based on the timestamp of the last spiked event (Fig.~\ref{fig:ats_vs_ts}). The resulting smooth gradient allows the optimizer to converge to the correct pose.

While time surfaces can create smooth gradients for pose estimation, choosing the correct decay rate is often challenging in real-world scenarios because the time surfaces may overlap or lose event data depending on camera motion speed and the complexity of the environment's textures if the decay rate is incorrectly set. \cite{manderscheid2019speed} proposed a speed-invariant time surface, but this approach only considers motion speed and does not account for texture complexity, which can be an issue in environments with rich textures. To address this limitation, we propose an adaptive time surface (ATS) that adjusts the decay rate based on both camera motion and environment textures. Our ATS computes the pixel-wise decay rate by analyzing the temporal event density of neighboring pixels. This allows the ATS to decay faster in regions experiencing high-texture environments or fast motion, generating a better-represented time surface map. Conversely, in parts of the ATS with low-texture environments or experiencing slow motion, the ATS keeps event data longer, resulting in clear and distinct selected pixels.

In addition to the challenges associated with time surfaces, detecting and maintaining distinctive pixels remain significant issues in direct methods too. For instance, \cite{zuo2022devo} selects pixels based on a constant threshold that can result in either too sparse or dense pixel selection, and the poorly distributed points can impede optimization~\cite{engel2017direct}. Also, picking pixels from distinctive features in the scene is important to achieve consistent key point matching since direct methods do not explicitly detect or track features. To address this issue, we propose a novel approach for selecting pixels directly from event data and performing filtering based on ATS. We employ two filtering processes. Firstly, we eliminate all pixels falling on the black areas of a median blurred ATS. Next, we further select only the pixels with high grayscale value and gradient of ATS. Additionally, we enforce a minimum distance between all selected pixels to achieve better point distribution. This approach helps to avoid selecting pixels in noisy regions and obtain well-distributed points in distinctive areas, such as the edges or corners of the scene.

\begin{figure}
\begin{center}
\includegraphics[width=1\linewidth]{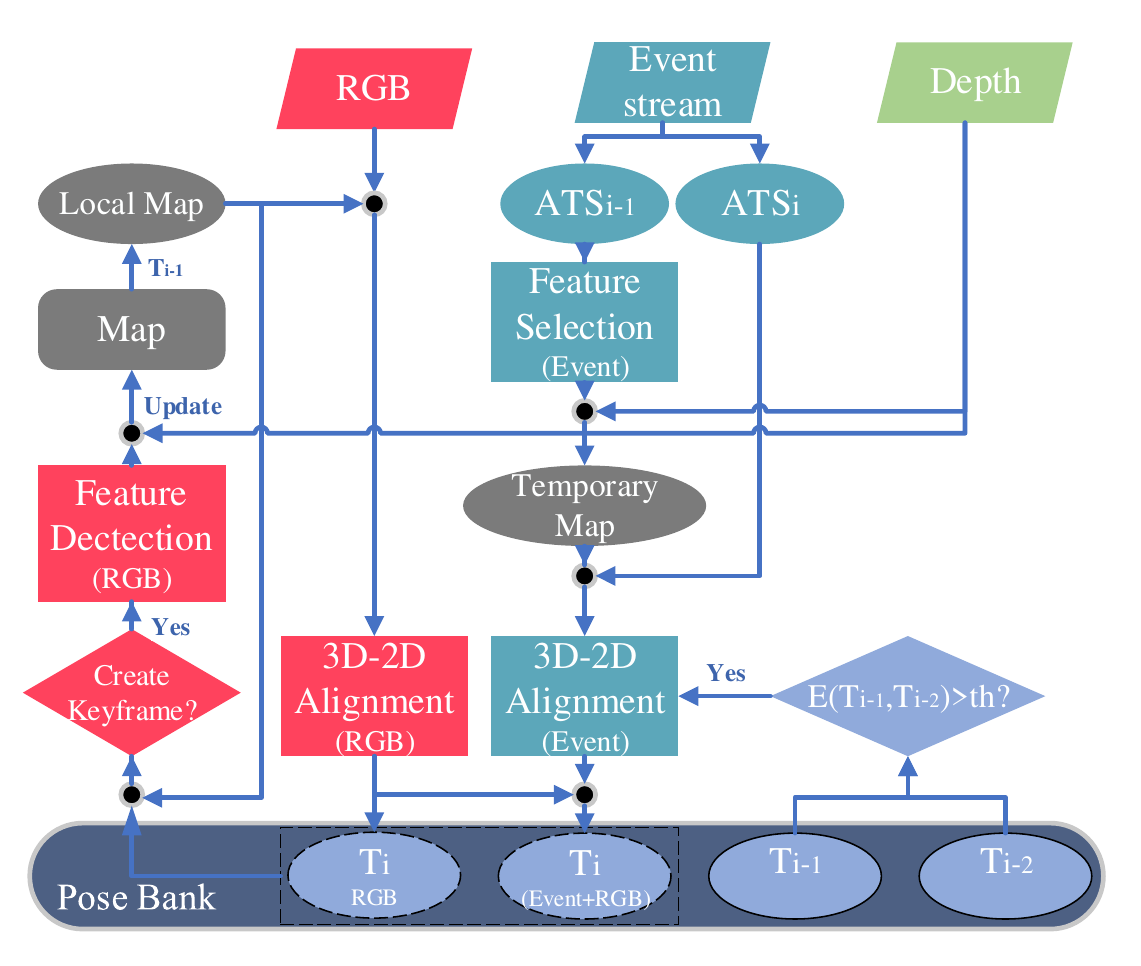}
\end{center}
\caption{{\bf Overview of our pose estimation framework.} During slow motion, we utilize RGB image-based pose estimation alone (indicated by red) to calculate the current pose, $\mathbf{T}_i$. When a substantial movement is detected, we activate the event stream for pose estimation by creating a temporary map and performing 3D-2D alignment (indicated by cyan) in conjunction with the RGB-based map and image.}
\label{fig:arch}
\end{figure}

Selected key points and correlated depth information are used to construct a map keyframe (Fig.~\ref{fig:arch}), or to compute a pose. During pose estimation, photometric errors in both RGB-based and event-based maps and images are simultaneously minimized to fully exploit both data streams. Through these three significant algorithmic improvements, namely ATS, pixel selection and filtering, and simultaneous optimization over RGB and event data, we achieved an accurate and robust pose estimator. Prior research on this topic has proposed an event-based direct method~\cite{hidalgo2022event}, but their algorithm may need to compromise the event camera's high temporal resolution because the event generation model (EGM) requires RGB images to generate a brightness increment image, which can be affected by the motion blur of RGB images.

Significant progress has been made in event-based visual odometry in recent years; however, prior methods have mainly been tested in aerial systems (e.g. drones) or wheeled ground vehicles, which typically do not undergo sudden changes in motion direction. Moreover, the primary purpose of pose estimation in aerial or wheeled systems has been obstacle avoidance or path following, which can tolerate relatively large errors in estimation accuracy. In contrast, legged robots traverse rough terrains by making contact with the ground, which demands greater robustness and accuracy of pose estimation. For example, even a slight error in estimation can cause stumbling or falling, leading to balance failure. Also, a dynamic maneuvering of a legged robot involves significant disturbance to vision sensors because of jerky movements and impacts from ground touchdowns. To the best of our knowledge, prior event camera-based estimation algorithms have not been tested on legged robots, and our experiments show that state-of-the-art event-based algorithms~\cite{vidal2018ultimate, qin2019a} quickly diverge once a robot makes dynamic locomotion involving aerial phase, highlighting the need for more robust and accurate pose estimation methods for legged robots.

Our contributions can be summarized as follows: 1) development of a direct-method-based estimation framework that integrates RGB-D and event data to achieve accurate and robust pose estimation of legged robots, without requiring an IMU sensor, 2) extensive algorithmic improvements of pixel selection and tracking, including a novel pixel-wise adaptive decay rate of time surface, an effective pixel selection algorithm using event data and our ATS, and a simultaneous pose estimation using both RGB-based and event-based data, and 3) compelling 6-DoF motion evaluations on both a public dataset and our own quadruped robot dataset. Our results demonstrate pose estimation less than $7~\si{\centi\meter}$ position error during dynamic locomotion such as trotting, pronking, and bounding. In addition, for the first time, our method successfully captures the acrobatic backflip motion of a quadruped robot (Fig.~\ref{fig:backflip}) without divergence.

\begin{figure}[t!]
    \centering\includegraphics[width=\linewidth]{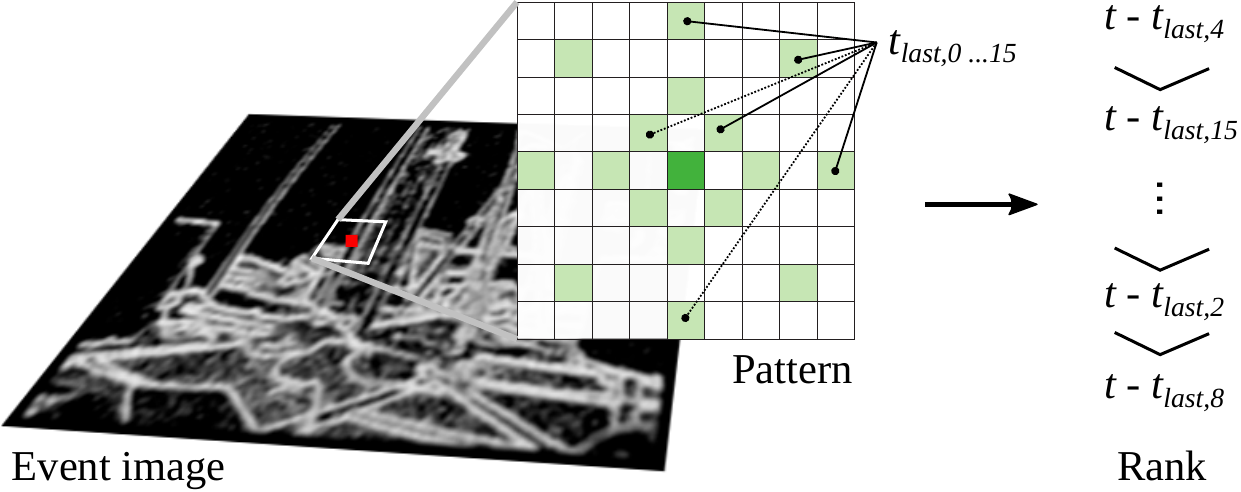}
    \caption{\textbf{Adjacent pixel selection pattern.} When building an ATS, we first pick 16 pixels (light green grids) around the target pixel(dark green grid) based on the given pattern. Among 16 pixels, we select the $n$ latest event data to compute the decay rate of the target pixel.}
    \label{fig:pixel_selection}
    \vspace{-0.3cm}
\end{figure}

\section{Pose Estimation Framework}
The core of the proposed method follows the PTAM~\cite{ptam} model that separates the SLAM system into tracking and mapping threads. As shown in Fig.~\ref{fig:arch}, the architecture takes synchronized and aligned RGB, depth, and event data as input. The left part is RGB and depth-based mapping (gray blocks) and tracking (red blocks) algorithm, which utilizes RGB-D data from depth images to construct a map based on an initial pose and estimated pose by a direct method using 3D-2D alignment. The right part shows a fusion strategy utilizing both RGB and event data (cyan blocks) to estimate a pose when a high-speed motion is detected. During normal operation, if the relative pose between consecutive frames is below a threshold, the tracking module employs only the RGB-based local map and the current RGB image in 3D-2D alignment to estimate the current pose $T_{i}^{\rm RGB}$. However, when the relative motion surpasses the threshold, the tracking module fuses information from the RGB-based local map, RGB image, event-based temporary map, and ATS to achieve superior tracking performance to estimate $T_{i}^{\rm RGB + Event}$. In the following sections, we will first explain the process of adaptive time surface map (\textbf{ATS}) construction and pixel selection, and then introduce the mapping and tracking modules.

\begin{figure}[t!]
    \centering\includegraphics[width=\linewidth]{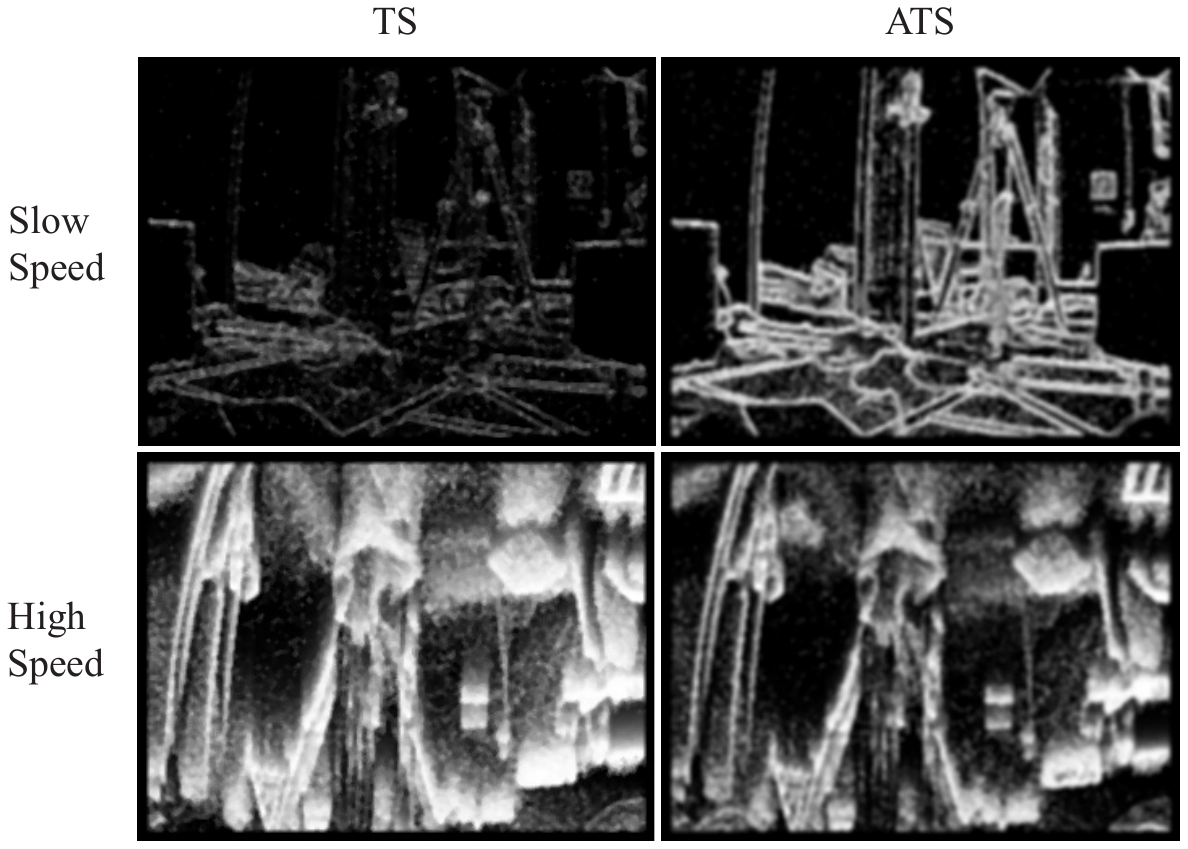}
    \caption{\textbf{Time surface (TS) and adaptive time surface (ATS).} Time surface and our adaptive time surface in walking motion (low speed) and backflip motion (high speed). ATS provides a clearer representation in both situations.}	
    \label{fig:ats_vs_ts}
    \vspace{-0.3cm}
\end{figure}

\subsection{Adaptive Time Surface}
A time surface map is a 2D image that visualizes the history of moving brightness patterns at each pixel and emphasizes the most recent event data with a higher grayscale value. Specifically, the grayscale value at each pixel location $\mathbf{x}$ is calculated based on the following equation:

% \begin{figure}[t!]
%     \centering\includegraphics[width=\linewidth]{img/system/Pattern_figure.pdf}
%     \caption{Candidate patterns to sample adjacent pixels. The Dark green pixel represents the center selected pixel. The light green pixels represent the adjacent pixels. Pattern 4 is used in this paper.}
%     \label{fig:pattern}
% \end{figure}

\begin{equation}
\mathcal{T}(\mathbf{x}, t)= 255 \times\exp \left(-\frac{t-t_{\text {last }}(\mathbf{x})}{\tau(\mathbf{x})}\right),
\end{equation}
where $\tau$ is typically set to a constant value, which makes all the pixels decay at the same ratio. However, depending on the camera motion and environment texture, the constant decay rate can cause an image with either too little or too much event data, neither of which is desirable. In this paper, we propose a novel adaptive time surface  that calculates pixel-wise decay rate, $\tau (\mathbf{x})$, based on the surrounding pixels' timestamp. The decay rate of ATS is calculated by
\begin{equation}
\tau(\mathbf{x}) = \max \left(\tau_u-\frac{1}{n} \sum_{i=0}^n (t - t_{last,i}), \tau_l\right),
\end{equation}
where $\tau_u$ and $\tau_l$ are the upper and lower bounds of the decay rate, respectively. $t_{last,i}$ is the timestamp of the $n$ latest pixels around $x$ that are selected by the patterns depicted in Fig~\ref{fig:pixel_selection}. Note that the pixels with $t_{last} = 0$, meaning that the pixels have not been activated, are not included in the ranking computation. Once we pick the $n$ latest pixels around $x$, then the upper bound is subtracted by the average of the time gap between the timestamp and the current time. The subtracted number sets the decay rate of the pixel $x$ unless it is smaller than the lower bound. Then blur and median blur filters are applied to produce a smoother result.

Our ATS utilizes this adaptive decay rate based on the complexity of the environment being processed. This approach ensures that the time surface decays faster in high-texture environments or during a high-speed motion to prevent overlapping pixels or white-out issues and construct a time surface with distinct pixels. In contrast, the ATS decays slower in low-texture environments or during low-speed motion, which ensures that the time surface captures sufficient information over a longer period, leading to improved pixel selection. Fig.~\ref{fig:ats_vs_ts} compares the ATS algorithm with a traditional time surface that uses a constant decay rate, $\tau$. The figure clearly demonstrates that the ATS algorithm produces a more distinct and clear surface under both slow-speed and high-speed motion.

\begin{figure}[t!]
    \centering\includegraphics[width=\linewidth]{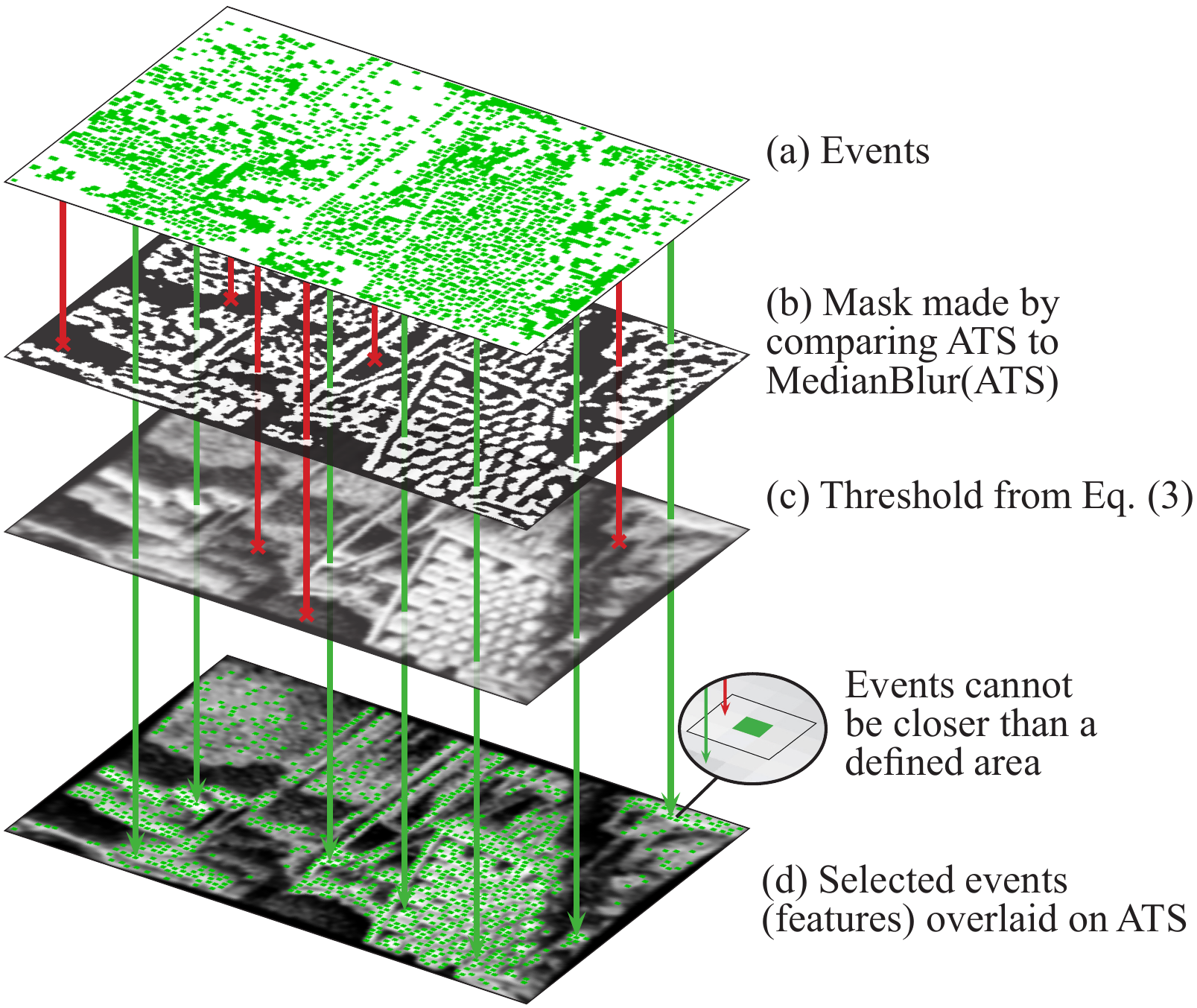}
    \caption{{\bf Pixel detection and filtering on ATS.} Events (a) are filtered through bright regions in the ATS through the use of a mask (b), then through a grayscale threshold (c). Each event must be a certain amount of pixels apart to prevent crowding. The result is a set of selected pixels (d). The rejected events are marked as red crosses at the end of red lines.}
    \label{fig:event_filtering}
    \vspace{-0.4cm}
\end{figure}

\subsection{Pixel selection and filtering}
We utilize different pixel selection strategies for RGB images and ATS. For RGB images, we followed the idea explained in \cite{engel2017direct} to ensure well-distributed key pixels. In our algorithm, we first divide an image into $d\times d$ blocks, then select pixels with gradients exceeding a certain threshold, which is adjusted based on the number of key pixels selected in each block. Therefore, each block has a different threshold depending on the underlying texture, and the adjusted thresholds help every block to contain an adequate number of key pixels.

In the case of pixel selection in ATS, we tried to avoid pure gradient-based methods to pick pixels from distinctive features in the scene. One issue of gradient-based key pixel selection in time surfaces (TS) is that TS usually employ filtering techniques that smooth out the gradients of TS, which can make it challenging to extract key pixels since high gradients are commonly used to identify them. To address this issue, we propose a novel pixel selection and filtering algorithm that selects pixels around the brightest regions in the ATS. Our approach starts by making a new image by applying a median blur to the original ATS, retaining only the pixels whose grayscale value on ATS image exceeds the median value of the blurred image, which is shown in Fig.~\ref{fig:event_filtering} (b). Next, we project events accumulated during a quarter duration of the time used for ATS building onto the mask, and only the points that fall on the white region of the mask are considered for further filtering. In the subsequent filtering round, we project the remaining events back to ATS, selecting only those with high grayscale value and gradient as final candidates, and the number is further reduced based on the distance between the points, which is summarized in Eq.~\eqref{eq:pixel}, and the resulting pixel selection and filtering output on the ATS is shown in Fig.~\ref{fig:event_filtering}(d).

\begin{equation}
\begin{split} \label{eq:pixel}
 S_{\text{pixel}}=\{(u, v) | &I(u, v)+\alpha \nabla_I(u, v)>h, \\ &\quad |u_i-u_j|>d, |v_i-v_j|>d \}
 \end{split}
\end{equation}
where $I(u,v)$ is the grayscale value and $\nabla_I(u, v)$ is the gradient for that pixel and $\alpha$ is a scale factor. New pixels should be away from the existing pixels for $d$ pixels both in X and Y coordinates.

\subsection{Mapping}
Fig.~\ref{fig:arch} illustrates the RGB and depth-based mapping module, which follows a conventional SLAM architecture. The mapping operation is only executed when inserting keyframes. We insert a keyframe based on the number of tracking pixels and their distribution. Specifically, the image is divided into nine regions, and each region is considered healthy if the number of tracking pixels exceeds a designated threshold. The total number of healthy regions and tracking pixels determines whether a keyframe should be inserted, new pixels selected, and new map points constructed. Additionally, each map point consists of multiple grayscale value arrays, with each array storing grayscale values of the new pixel and adjacent pixels on each pyramid layer.

To improve the tracking accuracy in dynamic environments, we construct a temporary map that is built when the relative motion between the previous two frames exceeds a predefined threshold. The temporary map utilizes detected pixels from the ATS and depth data as input, with each map point containing the same information as the primary map point, but with grayscale values sourced from ATS. We calculate the relative motion factor based on the angular velocity and linear velocity of the camera between the $T_{i-2}$ and $T_{i-1}$ frames. This factor can also be replaced by an IMU sensor or image blur detection module. The proposed approach significantly enhances the tracking robustness of our system in challenging, rapidly changing scenarios.

% \dhk{other options: Ephemeral map, Transitory map, instant map, On-demand map}
\begin{figure*}[t!]
\begin{center}
\includegraphics[width=1\linewidth]{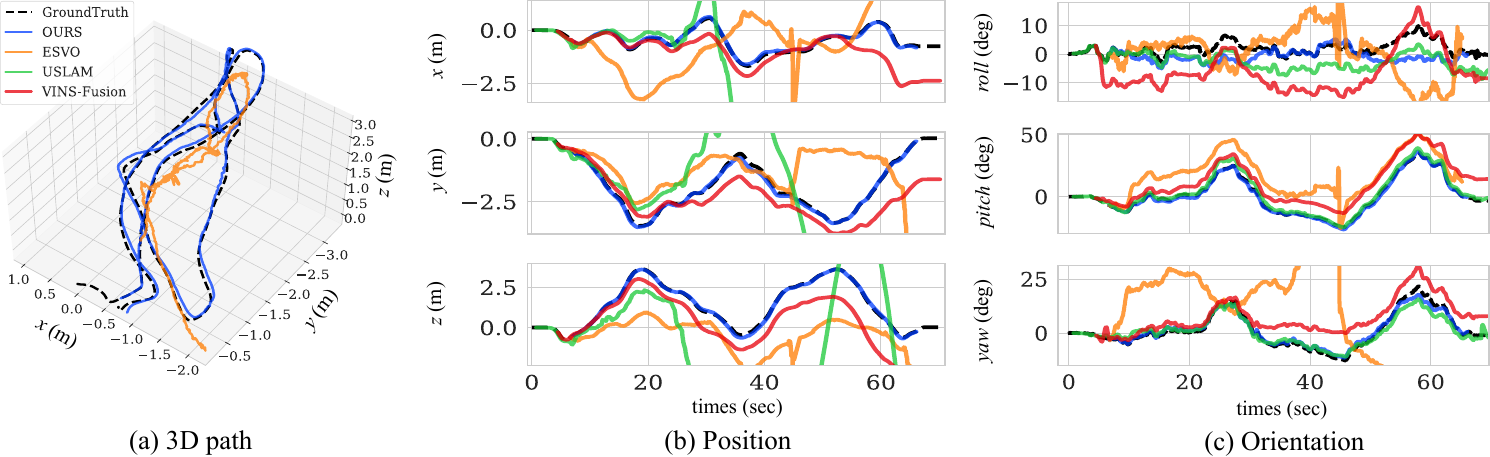}
\end{center}
\caption{ \textbf{Qualitative comparison of trajectory, position, and orientation of MVSEC indoor flying sequence} (a): the 3D paths of ESVO, GroundTruth, and our method. (b): Position error in XYZ axis. (c): Orientation in RPY order. (a) depicts the 3D path of the best accurate trajectory for clear representation. (b) and (c) are zoomed in to provide a closer view of the trajectory that is in proximity to the GroundTruth.}
\label{fig:mvsec_plot}
\vspace{0.3cm}
\end{figure*}

\begin{table*}[ht]
\setlength{\tabcolsep}{3pt}
\centering
\caption{Comparison on MVSEC Datasets. $\left[\mathbf{R}_{\text {rpe }}:~ ^{\circ}/ \mathrm{d}, \mathbf{t}_{\text {rpe }}:~ \mathrm{cm} / \mathrm{d}, \mathbf{t}_{\text {ate }}:~ \mathrm{cm}\right]$} \label{tab:mvsec_ours}
\begin{tabular}{lccclccclccclccc}
\hline & \multicolumn{3}{c}{ ESVO } & & \multicolumn{3}{c}{ VINS-Fusion } & &  \multicolumn{3}{c}{ USLAM } & & \multicolumn{3}{c}{ Ours }\\
\cline { 2 - 4 } \cline { 6 - 8 } \cline { 10-12 } \cline{ 14-16 }
 & 
$\mathbf{R}_{\text {rpe }}$ & $\mathbf{t}_{\text {rpe }}$ & $\mathbf{t}_{\text {ate }}$ & &
$\mathbf{R}_{\text {rpe }}$ & $\mathbf{t}_{\text {rpe }}$ & $\mathbf{t}_{\text {ate }}$ & &
$\mathbf{R}_{\text {rpe }}$ & $\mathbf{t}_{\text {rpe }}$ & $\mathbf{t}_{\text {ate }}$ & &
$\mathbf{R}_{\text {rpe }}$ & $\mathbf{t}_{\text {rpe }}$ & $\mathbf{t}_{\text {ate }}$\\
\hline flying1 & 
$0.69/1.02$ & $2.90/4.98$ & $14.40/47.65$ & &
$0.45$ & $2.38$ & $167.38$ & &
$0.42/0.45$ & $5.00/-$ & $58.56/-$ & &
$\textbf{0.36}$ & $\textbf{1.20}$ & $\textbf{6.89}$\\
flying2 & 
$0.96/-$ & $5.51/-$ & $338.28/-$ & &
$0.82/-$ & $7.07/-$ & $55.75/-$ & &
$0.59/-$ & $13.87/-$ & $159.22/-$ & &
$\textbf{0.49}$ & $\textbf{1.56}$ & $\textbf{8.92}$\\
flying3 & 
$0.59/0.64$ & $2.71/2.73$ & $11.10/16.03$ & &
$0.42$ & $2.77$ & $205.02$ & &
$0.39/-$ & $4.24/-$ & $31.27/-$ & &
$\textbf{0.36}$ & $\textbf{1.25}$ & $\textbf{5.26}$\\
flying4 & 
$-/-$ & $-/-$ & $-/-$ & &
$0.69$ & $3.79$ & $46.13$ & &
$\textbf{0.55}$ & $4.45$ & $38.58$ & &
$0.63$ & $\textbf{1.75}$ & $\textbf{9.16}$\\
\hline
\end{tabular}
\begin{tablenotes}
\item The symbol / separates evaluation on partial trajectory and full trajectory, where the partial trajectory is manually cut before the algorithm diverges significantly. The symbol -- indicates that the algorithm fails at an early stage of the experiment.
\end{tablenotes}
% \vspace{-0.3cm}
\end{table*}

\subsection{Tracking}
We have two tracking modules, one solely based on RGB images and another using event data along with the RGB images. The second module is activated when we detect large movement change, $\mathbf{T}_{i-1} - \mathbf{T}_{i-2} > threshold$. Here, we explain the second module, which integrates both RGB and event data. The primary goal of the tracking module is to find $T_i$ from the following equation,

\begin{equation}
\begin{split}\label{eq:pose_opt}
\min_{\mathbf{T}_i}&\quad \omega_1 \sum_{j \in M_{T_{i-1}}} e^{j^{\top}} W_p^j e^j +\omega_2 \sum_{k \in N_{T_{i-1}}} e^{k^{\top}} W_q^k e^k \\
\text{s.t.} & \\
& e^j=\sum_{p=0}^8\left(z_p^j-\pi_{\rm RGB}\left(\mathbf{T}_{i}^{\rm RGB} M_p^j\right)\right) \\
& e^k=\sum_{q=0}^{13}\left(z_q^k-\pi_{\rm event}\left(\mathbf{T}_{\rm RGB}^{\rm event} \mathbf{T}_{i}^{\rm RGB} N_q^k\right)\right),
\end{split}
\end{equation}
where $\omega_1$ and $\omega_2$ respectively are the weight factors for RGB-based tracking and ATS-based tracking, and $W_{p,q}$ are information matrices. $j$ is the index of visible map points given the previous pose, $\mathbf{T}_{i-1}$, in the RGB-based map $M$. $k$ denotes the index of the points in the event-based temporary map, $N$. $z^j_p$ are the saved grayscale values when we construct the map point $j$ (RGB), which are the same for $z_q^k$ and the map point $k$ (event). $p$ and $q$ are the indices of adjacent pixels around the selected pixel in the RGB-based map and event-based temporary map, respectively. The function $\pi$ represents the camera-to-image projection. In summary, Eq.~\eqref{eq:pose_opt} minimizes the errors between the saved grayscale values and the grayscale values of the image at the projected points from the maps to the image through the pose, $\mathbf{T}_i$. If we use an RGB-only estimation process, $\omega_2$ in Eq.~\eqref{eq:pose_opt} is set by zero.

The proposed strategy aims to mitigate the effects of motion blur while leveraging the constraints provided by RGB images and fusing event data together to provide better constraints. This complementary approach is advantageous because motion blur typically affects the parts of an RGB image that are perpendicular to the motion, and incurs pixel selection and tracking failures. In contrast, these regions often trigger the most events, which provide valuable constraints to the optimizer. By fusing event data with RGB images, the proposed approach can better leverage the strengths of each modality and provide more robust tracking results. In addition, constant motion and zero motion expectations are used to select the initial guess for RGB-based tracking. On the other hand, only zero motion prediction is applied to event-based tracking since the actual motion of the system is often random when event-based tracking is activated. 
% Experiments on RGB-only and fusion tracking are shown in Section \ref{Experiment}.

\section{Evaluation and Experiment Results} \label{Experiment}

We evaluate the performance of our estimation framework on a public dataset, called the Multi-Vehicle-Stereo-Event-Camera Dataset (MVSEC)~\cite{zihao2018multi}, and our self-collected dataset using a Mini-Cheetah robot. To ensure a fair comparison, several strategies were employed. Firstly, an $SE(3)$ alignment strategy is applied to the saved trajectory by taking the beginning frames into consideration. This is because each algorithm's local frame is defined when the algorithm is successfully initialized. Additionally, an $SO(3)$ alignment is applied to make the orientation of the first frame the same as the ground truth orientation. The alignment is done by EVO~\cite{grupp2017evo}. All the results are obtained by running the algorithms ourselves, except for DEVO~\cite{zuo2022devo}, where we directly adopt the accuracy results from their original paper as the source code is not available. For relative pose error, degrees per frame are compared with DEVO. And we choose degrees per degree as the evaluation metric to compare with other algorithms in our dataset because the dataset includes static motion.

Estimation results with an absolute trajectory error (ATE) greater than $5~\si{\meter}$ are considered as diverged, while relative pose errors (RPE) above $1~\si{\degree}/\text{d}$ (degree per degree) and $0.2~\si{\meter}/\text{d}$ (meter per degree) are also considered as diverged. If algorithms diverge in the middle of running, we compute the errors of the partial trajectory by cutting the trajectory before its estimation diverges. Note that no loop closure was performed to maintain consistency across all the algorithms.

% the beginning and end times may differ due to different initialization strategies and divergence. Moreover, 

\subsection{Experiment on MVSEC Dataset}
Four indoor sequences in MVSEC are used for the evaluation because they include synchronized event data, grayscale images, depth data obtained by a LiDAR, and ground-truth trajectories captured by a LiDAR-based algorithm, which are necessary to run various algorithms including ours. 
We have compared our method with four state-of-the-art algorithms:
\begin{enumerate}
    \item ESVO~\cite{zhou2021event}: A stereo visual odometry algorithm that utilizes two event cameras (input: stereo-event streams),
    \item DEVO: Latest event and depth data-based pose estimator (input: depth and event data), which has a similar sensor setup as our method,
    \item  UltimateSLAM (USLAM)~\cite{vidal2018ultimate}: state-of-the art event-based VIO that demonstrated great performance under aggressive motion (input: RGB images, event data, and IMU data),
    \item VINS-Fusion~\cite{qin2019a, vins_fusion_git}: a leading RGB and IMU-based VIO (input: RGB images from a mono camera and IMU data)
\end{enumerate}
%
% ESVO~\cite{zhou2021event}, which is a stereo visual odometry algorithm that utilizes event cameras only, DEVO, which shares similar setups with our method, UltimateSLAM (USLAM)~\cite{vidal2018ultimate}, which demonstrates great performance under aggressive motion, and VINS-Fusion~\cite{qin2019a}, which is a leading RGB and IMU-based method. 
%
All algorithms have been evaluated qualitatively and quantitatively on MVSEC. 
%To be more specific, ESVO utilizes stereo-event streams to produce visual odometry, while DEVO utilizes event and depth data. USLAM takes RGB images, event data, and IMU data as input, and VINS-Fusion utilizes RGB images from a mono camera and IMU data to produce odometry results.

\begin{table}[t!]
\setlength{\tabcolsep}{3pt}
\centering
\caption{Comparison on MVSEC Datasets. $\left[\mathbf{R}_{\text {rpe }}:~ ^{\circ}/ \mathrm{f}, \mathbf{t}_{\text {rpe }}:~ \mathrm{cm} / \mathrm{f}, \mathbf{t}_{\text {ate }}: ~\mathrm{cm}\right]$} 
\begin{tabular}{p{1.25cm } >{\centering\arraybackslash}p{0.7cm}>{\centering\arraybackslash}p{0.7cm}>{\centering\arraybackslash}p{0.7cm} p{0.35cm} >{\centering\arraybackslash}p{0.7cm}>{\centering\arraybackslash}p{0.7cm}>{\centering\arraybackslash}p{0.7cm}}
\hline & \multicolumn{3}{c}{ DEVO } & & \multicolumn{3}{c}{ Ours }\\
\cline { 2 - 4 } \cline { 6 - 8 }
 & 
$\mathbf{R}_{\text {rpe }}$ & $\mathbf{t}_{\text {rpe }}$ & $\mathbf{t}_{\text {ate }}$ & &
$\mathbf{R}_{\text {rpe }}$ & $\mathbf{t}_{\text {rpe }}$ & $\mathbf{t}_{\text {ate }}$\\
\hline flying1 & 
$0.30$ & $0.88$ & $20.58$ & &
$\textbf{0.15}$ & $\textbf{0.57}$ & $\textbf{6.89}$\\
flying2 & $0.36$ & $1.12$ & $11.33$ & &
$\textbf{0.20}$ & $\textbf{0.70}$ & $\textbf{8.92}$\\
flying3 & $0.53$ & $1.21$ & $10.60$ & &
$\textbf{0.15}$ & $\textbf{0.60}$ & $\textbf{5.26}$\\
flying4 & $0.53$ & $1.44$ & $13.16$ & &
$\textbf{0.26}$ & $\textbf{0.81}$ & $\textbf{9.16}$\\
\hline
\end{tabular}
% \vspace{-0.3cm}
\end{table}

Fig.~\ref{fig:mvsec_plot} shows the pose estimation results in terms of 3D trajectory, position, and orientation on indoor flying sequence data. In Fig.~\ref{fig:mvsec_plot}(a), only the ground truth, our proposed method, and partial ESVO trajectories are shown, as the position errors of the other two algorithms are clearly worse than ours, which can be found in the position plots (Fig.~\ref{fig:mvsec_plot}(b)). Table~\ref{tab:mvsec_ours} presents the quantitative results for all four indoor sequences, and both ATE and RPE are presented. Our method achieves less than $9.16~\si{\centi\meter}$ absolute trajectory error and outperforms all other algorithms in terms of both position and orientation. USLAM shows the best relative rotation error in flying4, which is a short and fast flying sequence where the IMU can provide accurate orientation information. Furthermore, algorithms that incorporate IMU sensor data are able to achieve decent orientation performance even when their ATE is large. 

One interesting finding of USLAM is that it can be initialized even when most of the camera's view is toward the ground, which does not contain many features. However, USLAM diverges in the middle of flying2 and flying3 datasets, as indicated by the $-$ symbol. This may be due to improperly performed feature detection because, when the feature distribution is poor, the condensed features in a small image region do not provide sufficient constraints to obtain a 6-DoF pose. VINS-Fusion shows good tracking performance, but the gradual drifting of position estimation leads to high absolute trajectory error. 

%VINS-Fusion requires the initial motion of IMU for initialization, making it unable to be initialized at the beginning. Our proposed method requires properly distributed features for initialization.

We consider ESVO to be the strongest competitor, as it achieves decent results on flying1 and flying3 datasets although it diverges in the middle of flying2 and flying4. As shown in Fig.~\ref{fig:mvsec_plot}(a), the trajectory exhibits significant vibration, which indicates unstable tracking. This can be attributed to the inability of the time surface to retain information over long periods, rendering tracking and triangulation vulnerable, particularly in slow motion.

%%%%%%%%%%%%%%%%%%%%%%%%%%%%%%%%%%%%%%%%%%%%%%%%%%%%%%%%%%%%%%%%%%%%%%%%%%%%
\begin{figure}[t!]
    \centering
    \includegraphics[width=\linewidth]{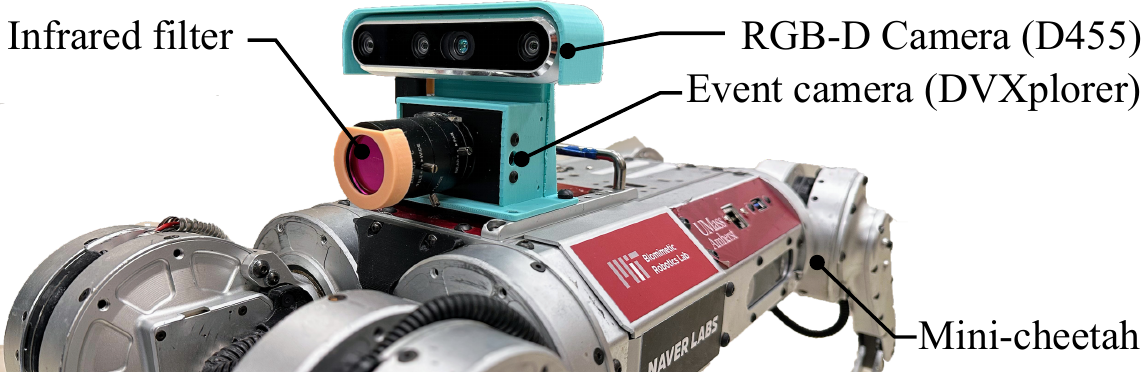}
    \caption{\textbf{Experimental setups.} Event camera and RGB-D camera are attached to the top of the Mini-cheetah robot. We use an infrared filter to filter out the infrared array spread by a depth camera.}	
    \label{fig:robot}
    \vspace{-0.3cm}
\end{figure}

\begin{figure*}[t!]
\begin{center}
\includegraphics[width=1\linewidth]{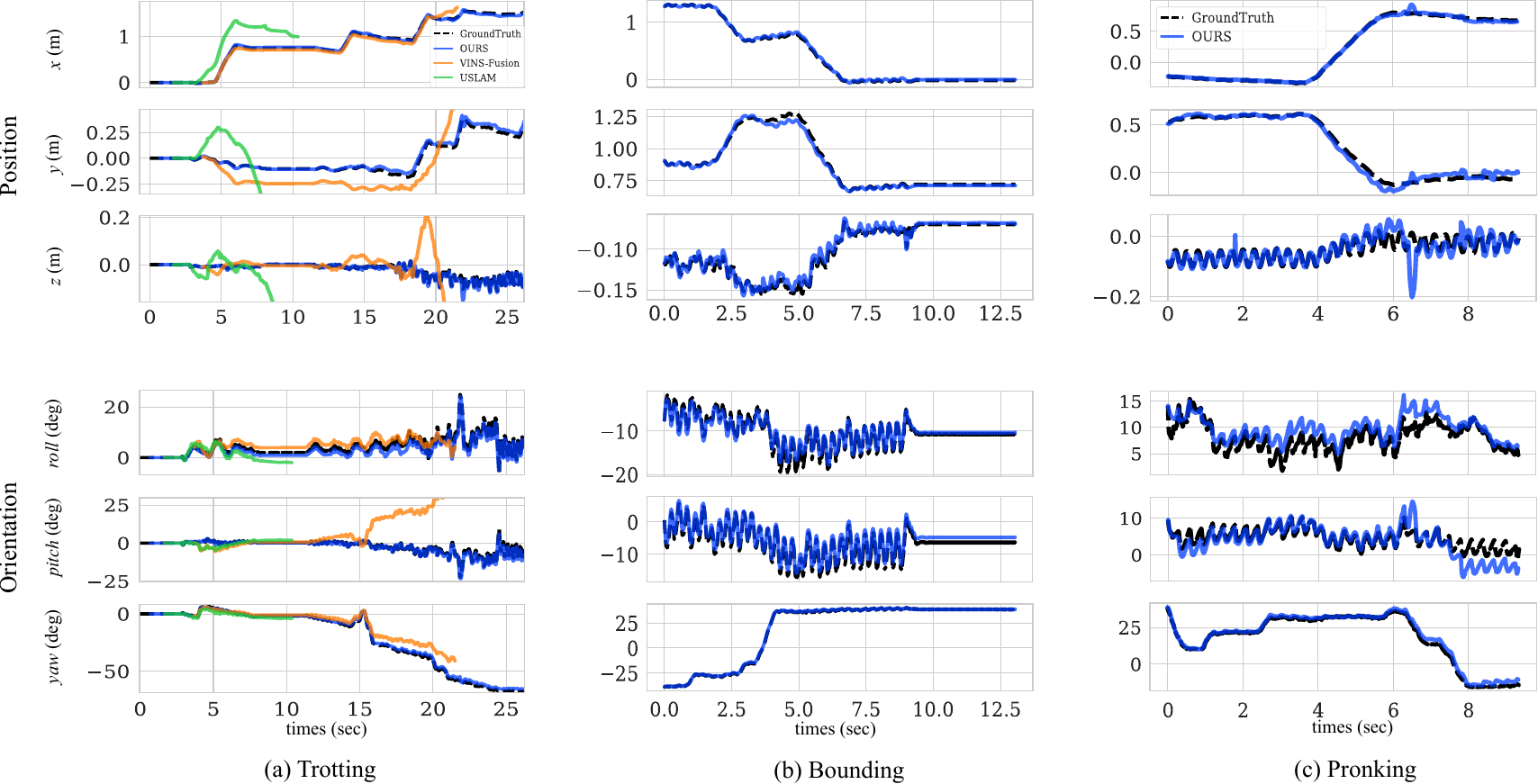}
\end{center}
\caption{\textbf{Estimation results during a sequence of trotting, bounding, and pronking motion.} The top and bottom rows present the position and orientation estimation, respectively. The ground truth is obtained by a motion capture system. (a) USLAM quickly diverges when the robot starts moving, while VINS-Fusion diverges when the robot's gait switches to bounding ($20~\si{\second}$), which is more dynamic than a trot gait. (b) The orientation change is more significant during the bounding. The proposed method accurately tracks the robot's pose even without IMU data. (c) The robot's motion along the $Z$-axis is large and involves significant impact disturbance during proking. However, our method is robust enough to maintain the stable position tracking of the robot.}
\label{fig:trot_bound_pronk}
\vspace{-0.3cm}
\end{figure*}

\subsection{Experiment on Our Quadruped Dataset}
We collect data from an event camera with $320 \times 240$ resolution (DVXplorer Lite)~\cite{inivation2020understanding} and a Realsense D455 camera mounted on top of the Mini-Cheetah robot as shown in Fig.~\ref{fig:robot}. The data was collected while the robot perform a range of dynamic motions, including trotting, pronking, bounding, and backflips. Notably, the backflip, pronking, and bounding motions involve significant aerial phase, with angular velocities up to $510~\si{\degree\per\second}$ in backflips and $260~\si{\degree\per\second}$ in bounding.

In the evaluation of pose estimation algorithms on our quadruped robot dataset, we compare the performance of our algorithm against VINS-Fusion and USLAM. In VINS-Fusion, we input RGB images and IMU data, while USLAM takes an event stream and IMU data as input since the event camera we are using does not have an RGB stream.

Note that our evaluation is conducted on a sequence of trotting, bounding, and pronking motions. Trotting is a relatively gentle walking gait, while bounding and pronking are more dynamic locomotion, as evidenced by the position and orientation change in Fig.~\ref{fig:trot_bound_pronk}. The experiment starts with trot gait and the gait is switched to bounding motion in $20~\si{\second}$. For a better understanding of the quadruped gaits, we refer to Fig.~5 in \cite{kim2019highly}.

In the experiment, we found that performance of USLAM, depicted by the green line in Fig.~\ref{fig:trot_bound_pronk}(a), significantly deteriorates as soon as the robot starts trotting. VINS-Fusion, on the other hand, maintains reasonable estimation accuracy during normal trotting but quickly diverges when the robot starts bounding. Our algorithm, however, demonstrates remarkable survivability and achieves an overall accuracy of $4.74~\si{\centi\meter}$ across all gaits. One potential reason for the failure of the other two algorithms is that the impact disturbance from touchdown is too large to maintain stable estimation because SLAM algorithms including USLAM and VINS-Fusion have been developed for wheeled robots or drones, which experience little impact disturbance during maneuvering.

\begin{table}[t!]
\centering
\setlength{\tabcolsep}{4.5pt}
\caption{Comparison on Our Recorded Datasets. $\left[\mathbf{R}_{\text {rpe }}:~ ^{\circ}/ \mathrm{d}, \mathbf{t}_{\text {rpe }}:~ \mathrm{cm} / \mathrm{d}, \mathbf{t}_{\text {ate }}:~ \mathrm{cm}\right]$}
\begin{tabular}{lccclccc}
\hline & \multicolumn{3}{c}{ VINS-Fusion } & & \multicolumn{3}{c}{ OURS }\\
\cline { 2 - 4 } \cline { 6 - 8 }
 & 
$\mathbf{R}_{\text {rpe }}$ & $\mathbf{t}_{\text {rpe }}$ & $\mathbf{t}_{\text {ate }}$ & &
$\mathbf{R}_{\text {rpe }}$ & $\mathbf{t}_{\text {rpe }}$ & $\mathbf{t}_{\text {ate }}$\\
\hline 
backflip1 &
$2.46/2.87$ & $4.17/-$ & $16.75/-$ & & $\textbf{1.45}$ & $\textbf{1.99}$ & $\textbf{8.51}$\\
backflip2 & 
$2.01/1.90$ & $2.60/-$ & $12.36/-$ & & $\textbf{1.18}$ & $\textbf{1.75}$ & $\textbf{5.31}$\\
running1 &
$0.82/-$ & $1.42/-$ & $22.08/-$ & & $\textbf{0.75}$ & $\textbf{0.55}$ & $\textbf{4.65}$\\
running2 & 
$1.01/-$ & $ 0.44/- $ & $15.51/-$ & & $\textbf{0.71}$ & $\textbf{0.56}$ & $\textbf{4.74}$\\
bounding & 
$-$ & $-$ & $-$ & & $\textbf{0.66}$ & $\textbf{0.42}$ & $\textbf{2.26}$\\
pronking & 
$-$ & $-$ & $-$ & & $\textbf{0.84}$ & $\textbf{0.92}$ & $\textbf{6.90}$\\
\hline
\end{tabular}
\begin{tablenotes}[flushleft]
\item running1 and running2 are a combination of different gaits, including trotting, bounding, pronking, etc. The symbol / separates evaluation on partial trajectory and full trajectory, where the partial trajectory is manually cut before the algorithm diverges significantly. The symbol -- indicates that the algorithm fails at an early stage of the experiment.
\end{tablenotes}
\vspace{-0.5cm}
\end{table}

% how to justify the omitted RGB data in USLAM?
USLAM wraps events onto the image plane and utilizes the IMU to perform motion compensation to obtain a sharper event image, which can be challenging for feature detection and tracking. The motion of legged robots contacting with the ground can be quite random, resulting in an inconsistent event image between adjacent frames. Moreover, contacting with the ground generates noisy IMU data, which renders the strategy of using IMU for motion compensation ineffective, particularly without the RGB input. 

VINS-Fusion succeeds in maintaining stable tracking during gentle trotting motion based on the RGB stream, but the more agile motion causes image blur, leading to feature tracking and pose estimation failure. In contrast to feature-based methods, the direct method that we use does not rely on feature detection and tracking. Instead, it utilizes all the available edge information in the image to provide constraints on the tracking module. This approach can be advantageous in situations where feature detection and tracking become challenging, such as in the case of aggressive motions where the images may become blurred. By using all the available edge information, the direct method can provide more robust pose estimation even in challenging scenarios. In addition, our algorithm takes advantage of both RGB-D data and event streams as input to constrain the tracking module. Specifically, the image blur caused by the aggressive motion lies on the texture of the image that is perpendicular to the motion, where many events are generated to provide constraints, making our algorithm effective during bounding and pronking. 
%Indeed, assuming zero motion or constant motion can also be beneficial for the tracking module, especially when the motion changes abruptly.

The proposed system is tested with an Intel Core i7-5820K CPU on a desktop computer. The proposed algorithm sequentially processes the data queue, with the overall optimization completed within $10~\si{\milli\second}$ and the RGBD-only tracking algorithm completed within $12~\si{\milli\second}$. In RGBD and event fusion mode, the tracking module takes approximately $80~\si{\milli\second}$. While the current running time is suboptimal, further improvements can be made to achieve real-time performance.

\section{Conclusions and Discussions}
We present a novel event camera-based visual odometry approach that utilizes both RGB-D and event data to enhance pose estimation accuracy. Our method incorporates a pixel-wise adaptive time surface generation strategy and efficient pixel selection method to provide more robust key points for the tracking module, particularly during aggressive motion. Our results demonstrate significant enhancement in accuracy and robustness over the sudden movements of a robot compared to prior visual odometry algorithms. We expect a meaningful extension of the legged robot application because of the improved pose estimation of agile systems, which has been a less highlighted and unsolved problem in the traditional SLAM domain. 

Due to limited access to codes and a compressed development timeline, we were unable to complete the evaluation of several recent algorithms~\cite{guan2022pl}\cite{hidalgo2022event}\cite{chen2022esvio} on our dataset. Continuous efforts will be made for further evaluation and comparison with other approaches in the future. Also, we plan to integrate an IMU sensor into the proposed method to exploit effectively its high-accuracy angular velocity measurement to the axes that are orthogonal to the gravity direction. 

\section*{ACKNOWLEDGMENT}
We express our gratitude to Naver Labs and MIT Biomimetic Robotics Lab for providing the Mini-cheetah robot as a research platform for conducting dynamic motion studies on legged robots.

% \addtolength{\textheight}{-12cm}   % This command serves to balance the column lengths on the last page of the document manually. It shortens  the textheight of the last page by a suitable amount. This command does not take effect until the next page it should come on the page before the last. Make sure that you do not shorten the textheight too much.

%%%%%%%%%%%%%%%%%%%%%%%%%%%%%%%%%%%%%%%%%%%%%%%%%%%%%%%%%%%%%%%%%%%%%%%%%%%%%%%%
% \section*{APPENDIX}

% \bibliographystyle{plainnat}
\bibliographystyle{IEEEtran}
\bibliography{main}

\end{document}